\documentclass[letterpaper, 10 pt, conference]{ieeeconf}  % Comment this line out if you need a4paper

\IEEEoverridecommandlockouts                              % This command is only needed if 
\usepackage{subcaption}

\usepackage{times}
\usepackage{epsfig}
\usepackage{graphicx}
\usepackage{amsmath}
\usepackage{amssymb}
\usepackage{booktabs}
\usepackage{array}
\usepackage{multirow}
\usepackage{makecell}
\usepackage{algorithm2e}
\usepackage[dvipsnames]{xcolor}
\RestyleAlgo{ruled}
\usepackage[pagebackref=true,breaklinks=true,letterpaper=true,colorlinks,bookmarks=false]{hyperref}
\usepackage[capitalize]{cleveref}
\crefname{section}{Sec.}{Secs.}
\Crefname{section}{Section}{Sections}
\Crefname{table}{Table}{Tables}
\crefname{table}{Tab.}{Tabs.}

\title{\LARGE \bf
Cross-Cluster Shifting for Efficient and Effective 3D Object Detection in Autonomous Driving
}

\author{Zhili Chen, Kien T. Pham, Maosheng Ye, Zhiqiang Shen, and Qifeng Chen% <-this % stops a space
\thanks{Zhili Chen (zchenei@connect.ust.hk), Kien T. Pham (tkpham@connect.ust.hk), Maosheng Ye (myeag@connect.ust.hk), and  Qifeng Chen (cqf@ust.hk) are with the Department of Computer Science and Engineering, HKUST. Zhiqiang Shen (Zhiqiang.Shen@mbzuai.ac.ae) is with the Department of Machine Learning, MBZUAI.}
}

\begin{document}

\maketitle
\thispagestyle{empty}
\pagestyle{empty}

\begin{abstract}
We present a new 3D point-based detector model, named Shift-SSD, for precise 3D object detection in autonomous driving. Traditional point-based 3D object detectors often employ architectures that rely on a progressive downsampling of points. While this method effectively reduces computational demands and increases receptive fields, it will compromise the preservation of crucial non-local information for accurate 3D object detection, especially in the complex driving scenarios. To address this, we introduce an intriguing Cross-Cluster Shifting operation to unleash the representation capacity of the point-based detector by efficiently modeling longer-range inter-dependency while including only a negligible overhead. Concretely, the Cross-Cluster Shifting operation enhances the conventional design by shifting partial channels from neighboring clusters, which enables richer interaction with non-local regions and thus enlarges the receptive field of clusters. We conduct extensive experiments on the KITTI, Waymo, and nuScenes datasets, and the results demonstrate the state-of-the-art performance of Shift-SSD in both detection accuracy and runtime efficiency.

\end{abstract}

\section{Introduction}
Object detection in the 3D point clouds plays a critical role in the fields of robotics and autonomous driving systems, allowing for accurate recognition and localization of objects. 
Given the sparseness and lacking topological information on point clouds, the existing approaches can be mainly categorized by their representations: point-based~\cite{shi2019pointrcnn,yang20203dssd,shi2020point,zhang2022not,yang2019std} and voxel-based approaches~\cite{zhou2018voxelnet,lang2019pointpillars,yin2021center,ye2020hvnet,deng2021voxel,kuang2020voxel,maturana2015voxnet}. Voxel-based approaches typically employ either hard or dynamic voxelization~\cite{zhou2020end} and then leverage the strengths of convolutional layers to enlarge the receptive fields rapidly. Though efficient and effective, the voxelization process inherently leads to a loss of geometry due to irreversible quantization. Additionally, the performance is considerably influenced by the voxel scale. In contrast, point-based approaches have demonstrated their ability to preserve the intrinsic geometry of point clouds and provide enhanced runtime efficiency~\cite{chen2022sasa,hu2020randla,qi2017pointnet++,yang20203dssd,zhang2022not}.

\begin{figure}[t!]
\centering
\includegraphics[width=0.48\textwidth]{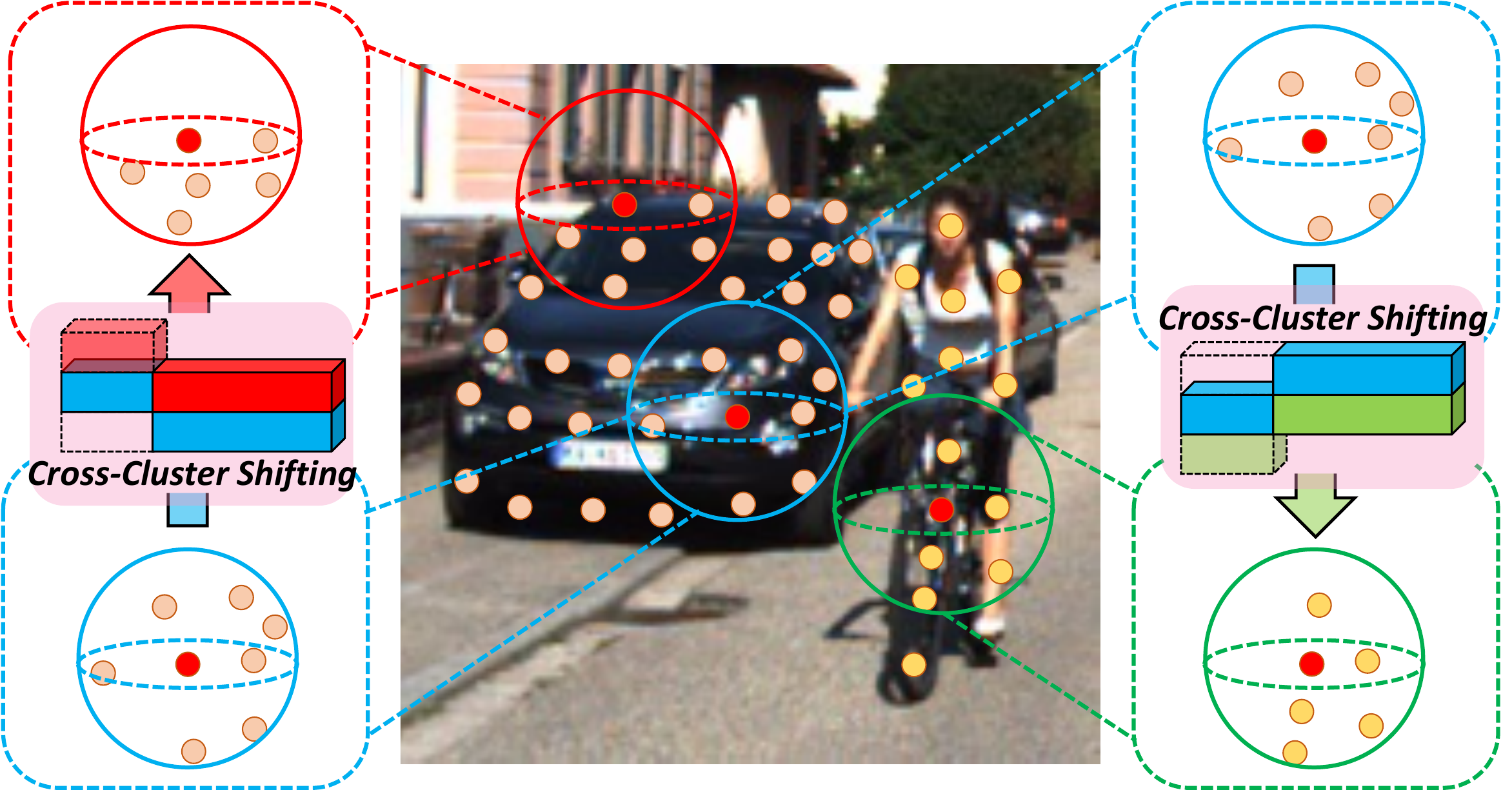}
\caption{3D point-based object detector commonly process point cloud data by first grouping points (denoted as \textcolor{orange}{orange}) around the selected cluster points (denoted as \textcolor{red}{red}) and then summarizing the local points' geometric patterns into the cluster points' features. Our proposed Shift-SSD builds interactions among the independently learned ball regions via Cross-Cluster Shifting. Shifting partial channels of the extracted features from \textcolor{blue}{blue} to \textcolor{red}{red} leads to better intra-instance learning, and from \textcolor{blue}{blue} to \textcolor{green}{green} resulting in more discriminative cross-instance learning.}
\label{fig:coreIdea}
\end{figure}

However, one challenge for point-based methods is balancing the degree of aggressive downsampling to save computational costs~\cite{fan2022fully} while not severely losing information for accurate predictions. The commonly employed set abstraction (SA) layers~\cite{qi2017pointnet++}, as well as their variations~\cite{li2018pointcnn,wu2019pointconv,ma2022rethinking,qian2022pointnext}, primarily focus on modeling local features for clustered points within a pre-defined spherical region. 

Unlike the traditional convolution operations that extract features through a sliding window, these methods struggle to effectively share information among point clusters grouped by balls, reminiscent of the constraints observed in sparse convolution~\cite{sun2022swformer}. Neglecting interaction between neighborhood cluster points leads to inferior feature propagation, further limiting the model representation capacity. The downsampling procedures further exacerbate the loss of information. Hybrid approaches~\cite{vora2020pointpainting,shi2021pv,shi2020pv,chen2017multi,zhou2020end} attempt to deal with this problem by introducing voxel representation to quickly enlarge the receptive fields with sparse convolution layers~\cite{yan2018second}. While these strategies have demonstrated some advancements, the additional memory and computational overhead remain significant concerns.

Motivated by the above analysis, we aim to attach more capability for feature interactions and information integration for point-based approaches. Inspired by the TSM~\cite{lin2019tsm} and ShuffleNet~\cite{zhang2018shufflenet}, we propose a novel point-based 3D detector, named \textbf{Shift-SSD}. Compared to TSM, which allows features to propagate along the temporal dimensions, and ShuffleNet, which helps information flow across channels, our proposed Cross-Cluster Shifting enables efficient feature propagation among cluster points. As illustrated in Fig.~\ref{fig:coreIdea}, features captured in different clusters extracted independently within grouped ball regions are exchanged and integrated to build long-range dependency, achieving the expansion of receptive fields. Compared with previous point-based approaches, our Shift-SSD armed with the proposed Cross-Cluster Shifting achieves remarkable improvement in 3D object detection, especially on the large-scale dataset regime. We conduct extensive experiments on the three datasets of KITTI Object Detection Benchmark~\cite{geiger2013vision}, Waymo~\cite{sun2020scalability}, and nuScenes~\cite{caesar2020nuscenes} datasets, and the reported superior performance justifies the effectiveness of our method.

In summary, our contributions reside as follows: 
\begin{itemize}
 \item We present an interesting information exchange scheme for 3D point-based object detectors, empowered with our simple yet effective {\em Cross-Cluster Shifting}. By efficiently modeling the correlation among the locally extracted features of clusters, the proposed {\em Cross-Cluster Shifting} expands the receptive fields with better information capture ability.

 \item We exhibit how the information exchange strategy boosts our proposed detector, and provide thorough analyses on our proposed {\em Cross-Cluster Shifting}.
 
 \item Extensive experiments on three datasets of KITTI~\cite{geiger2013vision}, Waymo~\cite{sun2020scalability}, and nuScenes~\cite{caesar2020nuscenes} demonstrate the superiority of our proposed {\em Shift-SSD}, in achieving state-of-the-art performance among existing point-based detectors while enjoying competitive inference speed.
\end{itemize}

\section{Related Work}
\noindent
\textbf{Point-based Detectors}
Point-based representation is the most straightforward way to represent a point cloud without the process of voxelization. PointRCNN~\cite{shi2019pointrcnn,chen2019fast} exploits voxel representation for initial bounding boxes generation and raw point clouds for second-stage refinement. Meanwhile, STD~\cite{yang2019std} conversely generates proposals from sparse point cloud input. These works follow a similar pipeline that first selects some farthest samples as cluster points to reduce the computational costs and apply PointNet++~\cite{qi2017pointnet++} or its variants~\cite{li2018pointcnn,le2018pointgrid,wu2019pointconv,qi2018frustum,wang2019frustum,cheng2021back,liu2019relation,wang2022rbgnet,2022dbqssd} as the backbone for predictions. Besides, 3DSSD~\cite{yang20203dssd} combines D-FPS with their proposed F-FPS to improve the quality of cluster point selections. IA-SSD~\cite{zhang2022not} and SASA~\cite{chen2022sasa} further incorporate their proposed learning-based class-aware sampling strategies with D-FPS for better efficiency. Our method also belongs to a point-based approach and focuses on enhancing point representation learning via the feature propagation procedure.

\begin{figure*}[t!]
\centering
\includegraphics[width=1.0\linewidth]{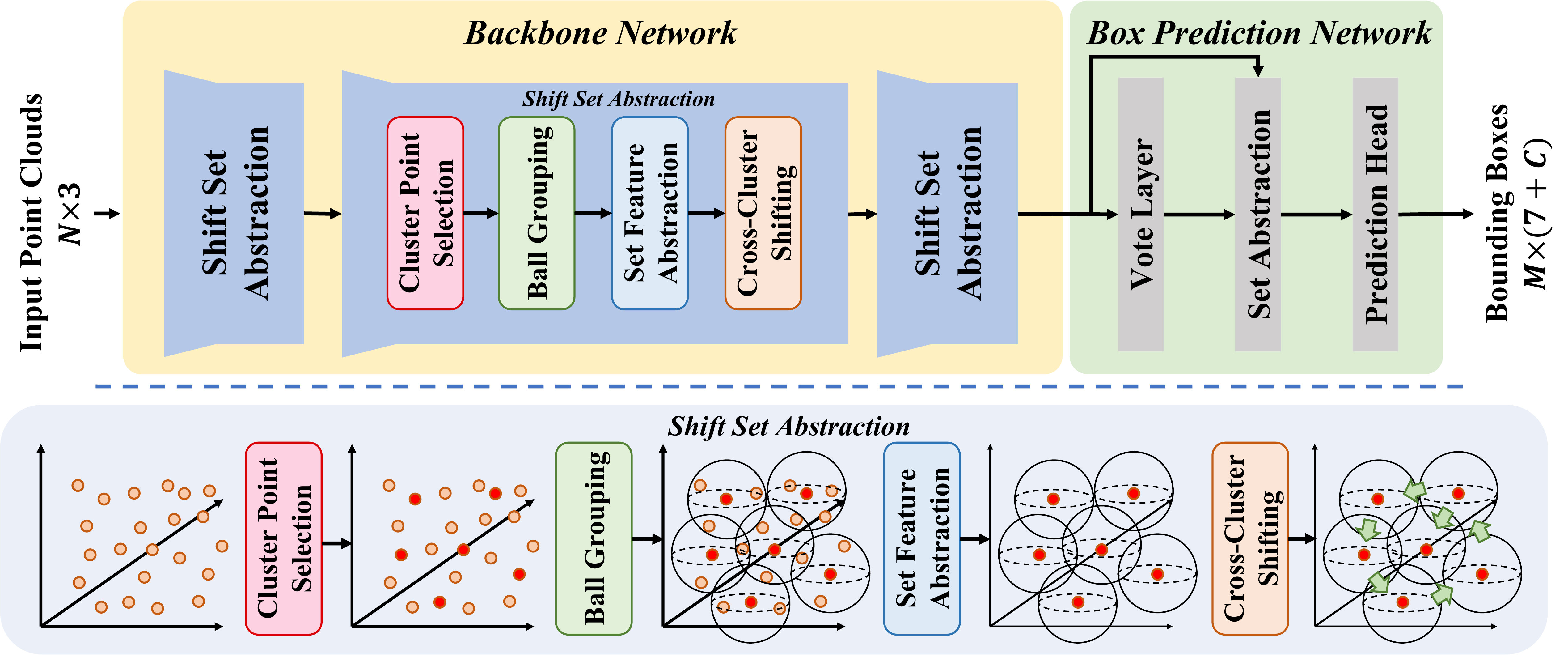}
\caption{The upper part of the figure presents the overall model architecture of the Shift-SSD and the detailed design of our SSA module. Shift-SSD comprises the Backbone Network and the Box Prediction Network. The Backbone Network takes raw point clouds as input and then conducts downsampling with a stack of our proposed SSA modules to summarize representative features into a point subset. As illustrated in the lower part of the figure, each SSA module applies Cluster Point Selection, Ball Grouping, and Set Featrure Abstraction to summarize local region features into cluster points. Followed by our proposed Cross-Cluster Shifting, it enhances the features by exchanging information among independently learned ball regions. The following Box Prediction Network first predicts offsets to shift cluster points towards instance centers with the Vote Layer~\cite{qi2019deep}, later using a Set Abstraction Layer to aggregate features. Lastly, the aggregated features are fed to the prediction head to generate bounding boxes with class labels.}
\label{fig:modelStructure}
\end{figure*}

\noindent
\textbf{Voxel-based Detectors}
Voxel representation is commonly used in 3D object detection since it converts the irregular data representation into a structural data format. With this, traditional convolution architecture can be directly applied for efficiency. VoxelNet~\cite{zhou2018voxelnet} is one of the pioneering works that combine learning-based voxel-wise feature extraction followed by dense 3D convolution. PointPillar~\cite{lang2019pointpillars} simplifies the dense 3D convolution with 2D convolution on the BEV space, which greatly improves the efficiency and saves the memory cost. Further, some extensions~\cite{ye2020hvnet,yan2018second,deng2021voxel,zheng2021se,yin2021center,chen2020object,qi2021offboard,ge20211st,hu2022afdetv2,yin2021centerpoint++,yang2022unified,zhou2022centerformer,he2022voxel,wang2022cagroup3d,dong2022mssvt,liu2023flatformer} are proposed based on the hierarchical feature learning to enhance the voxel-wise features with geometry guidance. Most of these works utilize sparse convolution~\cite{yan2018second} to improve the efficiency of the sparsity. Voxel-RCNN~\cite{deng2021voxel} and its variants~\cite{liang2020rangercnn} take the merits of two-stage frameworks to further refine the predictions. 

\noindent
\textbf{Hybrid Detectors}
Currently, more and more works pay attention to the fusion between different representations, including range views, point representation, and voxel representation. Works~\cite{shi2020pv,shi2021pv,ye2021drinet,tang2020searching,li2021voxel,li2021lidar,hu2022point,mao2021pyramid,chen2022mppnet,Fan2023SuperS3,Fan2023FSDVI} aim to integrate the merits of multiple representations. The general pipeline for this kind of work is to utilize sparse convolution~\cite{yan2018second,chen2022focal,Graham20173D} or convolution for voxel-wise feature extractions and refine the predictions with point-wise geometry learning. Point-wise and voxel-wise representations are simultaneously exploited for efficiency and performance. Moreover, transformer-based approaches~\cite{fan2022embracing,guan2022m3detr,mao2021voxel,wu2023virtual,chen2023futr3d,wang2023unitr} are further proposed for better cross-representation fusion. Compared with traditional attention layers, VoTr~\cite{guan2022m3detr} utilizes local attention and dilated attention mechanisms to capture the multi-scale context information at the sparse voxel level while alleviating the computation cost in the query process.
Exploring the direction of building a hybrid detector with our Shift-SSD is left for future work.

\section{Method}

\subsection{Overview}
The overall architecture of the proposed Shift-SSD is shown in Fig.~\ref{fig:modelStructure}. It consists of the backbone network for cluster point feature extraction and the box prediction network to produce detected boxes. The backbone network takes the raw point clouds as input and processes them with a stack of our proposed Shift Set Abstraction (SSA) modules to summarize features into a small subset of points. Then, the prediction network will first generate the candidate points by adding the predicted offsets to the cluster points and further aggregate features into the candidate points. Finally, the candidate points with the aggregated features will be sent to the regression and classification heads to predict the 3D bounding boxes with corresponding class labels.

\noindent\textbf{Backbone Network}
Several SSA modules are sequentially applied onto the input points to progressively downsample, achieving efficiency and producing point subsets (cluster points) with representative features. Each SSA comprises Cluster Point Selection, Ball Grouping, Set Feature Extraction, and Cross-Cluster Shifting layers.

\noindent\textbf{Box Prediction Network} The Box Prediction Network will first follow the VoteNet~\cite{qi2019deep} to generate the candidate points by predicting offsets that move the downsampled cluster points toward the instance centers. The candidate points are then considered as the selected cluster points to pass into a SA layer~\cite{qi2017pointnet++} to aggregate instance-level features for box prediction. The eventual aggregated instance features are sent to the regression head and classification head to predict with the 3D bounding boxes. The 3D bounding box proposals are post-processed by 3D NMS with a defined IoU threshold.

\subsection{Shift Set Abstraction}
\label{subsection:CCS}
Existing point-based 3D detection frameworks mostly inherit the PointNet series~\cite{qi2017pointnet,qi2017pointnet++} for processing the raw point cloud data. The intrinsic factor that prompts effective geometric feature learning for these frameworks is the flexible receptive field achieved by their proposed Set Abstraction (SA) layer. However, the original design of the SA layer only extracts features for cluster points within the ball regions under the predefined radii, and the learning processes of clusters are independent of each other within a layer, as illustrated in Fig.~\ref{fig:coreIdea}. We assimilate the sake of the SA layer and propose to leap forward with our proposed module, denoted as Shift Set Abstraction (SSA), to model longer-range dependencies by diffusing local geometric information among clusters with the operation of Cross-Cluster Shifting, as presented in the lower part of Fig.~\ref{fig:modelStructure}.

\noindent\textbf{Set Abstraction}
Given a set of input points $\mathcal{P}=\{p_1,...,p_n\}$ for each SA layer, the Cluster Points Selection utilizes the sampling operation of D-FPS~\cite{qi2017pointnet} or Ctr-S~\cite{zhang2022not} to obtain the representative cluster points. To capture local geometric patterns and further summarize them to cluster points, Ball Grouping is first applied to sample the neighbors for each cluster point. Then the Set Feature Abstraction, consisting of an MLP layer and a reduction operation, is conducted within each independently grouped region to summarize local geometric features. These summarized features of each cluster point are considered the information carrier of their representative ball regions. We obtain multi-scale features for cluster points by applying the similar aforementioned process through setting different radii for the Ball Grouping. We denote the summarized cluster features extracted under the radius of $r$ as $\mathbf{x}^{r}_i$. The features that summarize under different radii will be concatenated and then fusion with an MLP. The Set Abstraction can be formulated as
\begin{align}
\label{eqn:1}
\mathbf{x}^{r}_i &= \mathcal{R}(\mathcal{F}([\mathbf{x}_k, p_k-p_i])\rvert k = 1,...,K),\\
\mathbf{x}^{a}_i &= \mathcal{A}([\mathbf{x}^{r}_i\rvert r = 1,...,R]),
\label{eqn:2}
\end{align}
where $p_i$ represents the cluster point. $p_k$ is the neighbor point of $p_i$ and is sampled within the ball region, which is defined by radius $r$. $\mathbf{x}_k$ are the features of neighbor point $p_k$. $\mathcal{F}$ represents the $\mathbf{MLP}$ for extracting features under the scale $r$, which takes the concatenation of $\mathbf{x}_k$ and the relative coordinate of $p_k-p_i$ as input and output with neighbor features. Later the neighbor features are summarized with the reduction layer $\mathcal{R}$ (max-pooling) as the cluster's local geometric features, denoted as $\mathbf{x}^{r}_i$. As shown in Eqn.~\ref{eqn:2}, the summarized cluster features $\mathbf{x}^{r}_i$ of different scales are concatenated and further fused by an aggregation layer (an $\mathbf{MLP}$ denoted as $\mathcal{A}$).

\noindent\textbf{Motivation}
As illustrated in Eqn.~\ref{eqn:1}, we observed the cluster point features $x^r_i$ are independently extracted within a limit ball region constrained by $r$. The prior work of ShuffleNet~\cite{zhang2018shufflenet} helps the information flow among channel groups with the channel shuffle operation and TSM~\cite{lin2019tsm} enables temporal modeling by shifting channels across frames. Motivated by the observed limitation in the traditional design and the existing works, we introduce the novel Cross-Cluster Shifting to the Set Abstraction, which actualizes the inter-dependency modeling among those independently learned cluster points, as shown in the lower part of Fig.~\ref{fig:modelStructure}.

\begin{figure}[t!]
\centering
\includegraphics[width=1.0\linewidth]{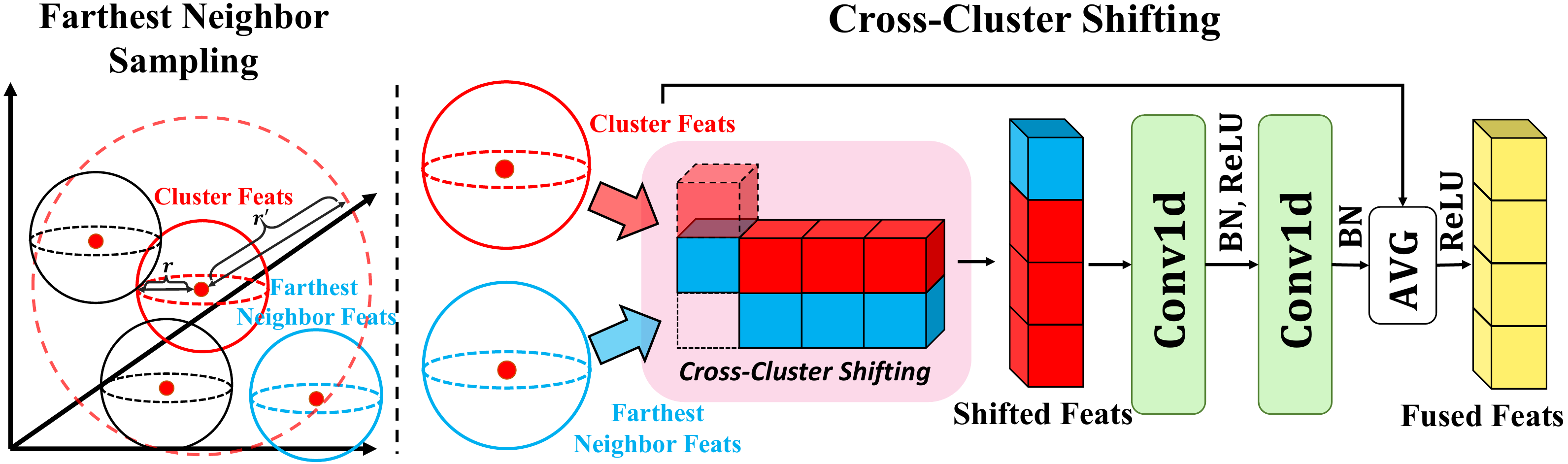}

\caption{The pipeline of the Cross-Cluster Shifting. The cluster features in the center is in \textcolor{red}{red}. As shown on the left of the figure, we first utilize the farthest neighbor sampling to obtain its farthest neighbor in \textcolor{blue}{blue} within the range of $r^\prime$. Then, Cross-Cluster Shifting is conducted to exchange partial features from the farthest neighbor to the cluster features. The resulting fused features in \textcolor{yellow}{yellow} are obtained by passing through two Conv layers, followed by an average pooling operation.}
\label{fig:CSU}

\end{figure}

\noindent\textbf{Farthest Neighbor Sampling} We design the Cross-Cluster Shifting to conduct interaction for the cluster points $p_i$ with each of their farthest neighbors denoted as $p_f$ to incorporate more information from their surrounding environment. Given the cluster points are downsampled through D-FPS~\cite{qi2017pointnet} or Ctr-S~\cite{zhang2022not}, they are sparser (spatially farther from each other). Therefore, the range defined by radius $r^\prime$ for searching the farthest neighbors should be larger in order to form a diverse neighbor cluster points set and later pick the farthest one from it. As shown on the left of Fig.~\ref{fig:CSU}, we utilize the ball-query operation, the same as the one conducted in Ball Grouping~\cite{qi2017pointnet++}, by first randomly sampling $K$ neighbors within a larger range of $r^\prime$ for each cluster point and then picking the farthest one among these sampled neighbors. The pairing between each cluster and its farthest neighbor is shared across the multi-scale branches. Note that the farthest neighbor sampling is defined with the same radius of $r^\prime$ as the one used in the later SSA module, such that the formed neighbor cluster points set can be reused for saving computation. We analyze the effectiveness of the selection strategy that picks the farthest neighbor of each cluster point to interact with in Section~\ref{section:expr}.

\noindent\textbf{Cross-Cluster Shifting}
To propagate the cluster-level information between the cluster and its farthest neighbor, we reserve an information-shared features space with $s$ channels for each cluster point feature $\mathbf{x}^{r}_i$ to shift in the information stored in its farthest neighbor $\mathbf{x}^{r}_f$. As shown at the right of Fig.~\ref{fig:CSU}, we first apply the gather operation to obtain the features with $s$ channels from the farthest neighbors corresponding to each cluster point. Then, we shift these features from the farthest neighbors toward the information-shared region of each cluster point. We further pass these features into a two-layer MLP to interact the shift-in features with the remaining features that store the local geometric information, as illustrated in Fig.~\ref{fig:CSU}. To maintain the local geometric feature learning capacity of the cluster point, we insert the Cross-Cluster Shifting into the residual branch followed by taking the average on the resulting features with the input features of ${x}^{r}_i$. The Cross-Cluster Shifting is formed as follows:
\begin{align}
\mathbf{h}^{r}_i &= \textbf{AVG}(\mathbf{MLP}([\mathbf{x}^{r}_f[:s],\mathbf{x}^{r}_i[s:]]),\mathbf{x}^{r}_i),
\label{eqn:3}
\end{align}
where $\mathbf{h}^{r}_i$ are the enhanced cluster features output from the Cross-Cluster Shifting. $\mathbf{h}^{r}_i$ will then be activated by $\mathbf{ReLU}$ after Eqn.~\ref{eqn:3}.

We rewrite the aggregation layer defined in Eqn.~\ref{eqn:2} as
\begin{equation}
\mathbf{h}^{a}_i = \mathcal{A}([\mathbf{h}^{r}_i\rvert r = 1,...,R]),
\label{eqn:4}
\end{equation}
where $\mathbf{h}^{a}_i$ is the input for the next Shift Set Abstraction layer.

\subsection{End-to-end Learning}
We train Shift-SSD in an end-to-end training manner. The overall loss function consists of centroid offset loss $L_{offset}$, classification loss $L_{cls}$, and box regression loss $L_{box}$:
\begin{equation}
\label{eqn:9}
L = \lambda_1 L_{offset} + \lambda_2 L_{cls} + \lambda_3 L_{box}.
\end{equation}
$L_{offset}$, is calculated by smooth-$L1$ to supervise the Vote layer~\cite{qi2019deep} to regress the clusters' offsets towards the instance centroids. We utilize cross-entropy loss in $L_{cls}$ for training the classification of bounding boxes. We apply the box regression loss $L_{box}$ as ~\cite{yang20203dssd, zhang2022not}, which constitutes losses for regressing location, box size, angle, and distance to corner:
\begin{equation}
\label{eqn:10}
L_{box} = \delta_1 L_{loc} + \delta_2 L_{size} + \delta_3 L_{angle} + \delta_4 L_{corner}.
\end{equation}
For the hyper-parameters of different losses, we follow ~\cite{yang20203dssd, zhang2022not} to set $[\lambda_i]_{i\in\{1,2,3\}}$ and $[\delta_j]_{j\in\{1,2,3,4\}}$ as $1.0$.
   \vspace{0.1cm}
\begin{table*}[t!]
   \caption{Quantitative comparison on the Waymo \textit{validation} set. Our results are shown in bold, and the best results of each category are \underline{underlined}.}
   \vspace{-0.3cm}
   \begin{center}
      \renewcommand{\arraystretch}{1.05}
   \resizebox{1.0\textwidth}{!}{  

   \begin{tabular}{r|c c|c c|c c|c c|c c|c c}
\Xhline{2.0\arrayrulewidth}
   \multirow{2}{*}{Method} & \multicolumn{2}{c|}{Vehicle (LEVEL 1)} &\multicolumn{2}{c|}{Vehicle (LEVEL 2)} & \multicolumn{2}{c|}{Ped. (LEVEL 1)} & \multicolumn{2}{c|}{Ped. (LEVEL 2)} & \multicolumn{2}{c|}{Cyc. (LEVEL 1)} & \multicolumn{2}{c}{Cyc. (LEVEL 2)} \\
   & mAP & mAPH & mAP & mAPH & mAP & mAPH & mAP & mAPH & mAP & mAPH & mAP & mAPH \\

\Xhline{2.0\arrayrulewidth}
    PointPillars \cite{lang2019pointpillars} & 60.67 & 59.79 & 52.78 & 52.01 & 43.49 & 23.51 & 37.32 & 20.17 & 35.94 & 28.34 & 34.60 & 27.29 \\
    SECOND \cite{yan2018second} & 68.03 & 67.44 & 59.57 & 59.04 & 61.14 & 50.33 & 53.00 & 43.56 & 54.66 & 53.31 & 52.67 & 51.37 \\ 
    Part-$A^2$ \cite{shi2020points} & 71.82 & 71.29 & 64.33 & 63.82 & 63.15 & 54.96 & 54.24 & 47.11 & 65.23 & 63.92 & 62.61 & 61.35 \\ 
    PV-RCNN \cite{shi2020pv} & 74.06 & 73.38 & 64.99 & 64.38 & 62.66 & 52.68 & 53.80 & 45.14 & 63.32 & 61.71 & 60.72 & 59.18 \\ 
    IA-SSD \cite{zhang2022not} & 70.53 & 69.67 & 61.55 & 60.80 & 69.38 & 58.47 & 60.30 & 50.73 & 67.67 & 65.30 & 64.98 & 62.71 \\
    DBQ-SSD \cite{2022dbqssd} & 71.58 & 71.03 & 64.13 & 63.61 & 69.18 & 58.47 & 60.22 & 50.81 & \underline{68.29} & 66.01 & \underline{66.09} & 63.86 \\
    \cline{1-13}
    \textbf{Shift-SSD (Ours)} & \underline{\textbf{74.15}} & \underline{\textbf{73.6}} & \underline{\textbf{65.1}} & \underline{\textbf{64.6}} & \underline{\textbf{72.36}} & \underline{\textbf{62.31}} & \underline{\textbf{63.41}} & \underline{\textbf{54.53}} & \textbf{68.24} & \underline{\textbf{66.42}} & \textbf{66.06} & \underline{\textbf{64.29}} \\
\Xhline{2.0\arrayrulewidth}
   \end{tabular}}
   \end{center}
   \label{tab:waymo_val}
  \vspace{-0.2cm}
   \end{table*}

\begin{table*}[t!]
   \caption{Quantitative comparison on the nuScenes \textit{validation} set. Our results are shown in \textbf{bold}, and the best results of each category are \underline{underlined}. $\dagger$ denotes results derived from~\cite{yang20203dssd} and $\ast$ indicates training conducted by us.}
      \vspace{-0.3cm}
   \begin{center}
    \renewcommand{\arraystretch}{1.05}
   \resizebox{\textwidth}{!}{  

   \begin{tabular}{r|c|c|c|c|c|c|c|c|c|c|c}
\Xhline{2.0\arrayrulewidth}
    \multirow{1}{*}{Method}&  \multirow{1}{*}{mAP} & \multirow{1}{*}{Car}& \multirow{1}{*}{Ped.}& \multirow{1}{*}{Bus}& \multirow{1}{*}{Barrier}& \multirow{1}{*}{TC}& 
    \multirow{1}{*}{Truck}& \multirow{1}{*}{Trailer}&
    \multirow{1}{*}{Motor}& \multirow{1}{*}{Cons. Veh.}&
    \multirow{1}{*}{Bicycle}\\

\Xhline{2.0\arrayrulewidth}

   \multirow{1}{*}
   SECOND$^\dagger$ \cite{yan2018second} & 27.12 & 75.53 &59.86& 29.04& 32.21& 22.49& 21.88& 12.96& 16.89& 0.36& 0.0  \\
   \multirow{1}{*} PointPillars$^\dagger$ \cite{lang2019pointpillars}  & 29.5 & 70.5 &59.9& 34.4& 33.2& 29.6& 25.0& 20.0& 16.7& 4.5& 1.6\\

   \multirow{1}{*} 3DSSD \cite{yang20203dssd} & 42.66 & \underline{81.20}& \underline{70.17}& 61.41& 47.94& \underline{31.06}& \underline{47.15}& 30.45& 35.96& 12.64& 8.63\\
   \multirow{1}{*} IA-SSD$^\ast$~\cite{zhang2022not} &	42.23 & 71.91	&64.36	&66.90&	48.40& 29.23& 45.49&31.50&	34.74& 15.36&	14.32\\
   \cline{1-12}

    \textbf{ Shift-SSD (ours) }  &  \underline{\textbf{44.39}} &\textbf{72.64} & \textbf{68.80} & \underline{\textbf{67.79}} & \underline{\textbf{51.13}} & \textbf{30.32} & \textbf{46.86} & \underline{\textbf{34.01}} & \underline{\textbf{37.73}} & \underline{\textbf{17.69}} & \underline{\textbf{16.92}}\\

\Xhline{2.0\arrayrulewidth}
   \end{tabular}}
   \end{center}

   \label{tab:nuscenesvalmAP}
       \vspace{-0.2cm}
\end{table*}

\section{Experiments}
\label{section:expr}
We conduct experiments to evaluate our model using three well-known benchmarks: the KITTI~\cite{geiger2013vision}, Waymo~\cite{sun2020scalability}, and nuScenes~\cite{caesar2020nuscenes} datasets.

\begin{table}[h]
   \caption{Quantitative comparison on the KITTI \textit{test} set at Moderate level. All results are evaluated via the oﬀicial evaluation server. Our results are shown in \textbf{bold}, and the best results of each category are \underline{underlined}. 
    \vspace{-0.3cm}
   \label{tab:kittitest}}
    \begin{center}
   \resizebox{0.475\textwidth}{!}{
      \renewcommand{\arraystretch}{1.05}
      \begin{tabular}{c|r|p{14mm}<{\centering}|p{14mm}<{\centering}|p{14mm}<{\centering}@{}}
      \Xhline{2.0\arrayrulewidth}
       \multirow{2}{*}{}&\multirow{2}{*}{Method} & \small{Car Mod.} & \small{Ped. Mod.} & \small{Cyc. Mod.}  \\

       & & (IoU=0.7) & (IoU=0.5) & (IoU=0.5)  \\
      \Xhline{2.0\arrayrulewidth}

   \multirow{5}{*}{\rotatebox{90}{Voxel-based}}
   & SECOND \cite{yan2018second} & 75.96 & 35.52 & 60.82  \\

   & PointPillars \cite{lang2019pointpillars} & 74.31 & 41.92 & 58.65 \\

   & SA-SSD \cite{he2020structure} & 79.79 & - & - \\ 

   & TANet \cite{liu2020tanet} & 75.94 & \underline{44.34} & 59.44 \\  
   & Part-$A^2$ \cite{shi2020points} & 78.49 & 43.35 & 63.52 \\

   \hline
   \multirow{4}{*}{\rotatebox{90}{Point-Voxel}}
   & STD \cite{yang2019std} & 79.71 & 42.47 & 61.59 \\

   & PV-RCNN \cite{shi2020pv} & 81.43 & 43.29 & 63.71 \\ 
   & HVPR \cite{noh2021hvpr} & 77.92 & 43.96 & - \\ 
   & VIC-Net \cite{jiangty2021vicnet} & 80.61 & 37.18 & 63.65 \\ 
   \hline
   \multirow{5}{*}{\rotatebox{90}{Point-based}} 
   & PointRCNN \cite{shi2019pointrcnn} & 75.64 & 39.37 & 58.82 \\ 

   & 3DSSD \cite{yang20203dssd} & 79.57 & 44.27 & \underline{64.10} \\

   & IA-SSD \cite{zhang2022not}  & 80.13 & 39.03 & 61.94 \\
   & DBQ-SSD \cite{2022dbqssd}  &  79.39 & 38.08 & 62.80 \\

   \cline{2-5}

   & \textbf{ Shift-SSD (Ours)} & \underline{\textbf{81.65}} & \textbf{36.74} & \textbf{63.13}\\

\Xhline{2.0\arrayrulewidth}
   \end{tabular}}

   \end{center}
    \vspace{-0.55cm}
\end{table}

\subsection{Implementation Details and Results}
\noindent
\textbf{KITTI}
dataset provides 80$\mathrm{K}$ labeled 3D objects over 15$\mathrm{K}$ LiDAR samples. Following the predecessors, we employ the same protocol to preprocess the point cloud data before training. Subsequently, we adopt settings on network depth and width for our Shift-SSD similar to ~\cite{qi2017pointnet++, zhang2022not, yang20203dssd, 2022dbqssd} by stacking our SSA modules to sequentially downsample input point clouds to $4096$ $\rightarrow$ $1024$ $\rightarrow$ $512$ $\rightarrow$ $256$ points, and simultaneously extract point-wise features under two different scales. The shifting ratio is empirically set as $1/8$, the best one we
select from \{$1/16$, $1/8$, $1/4$, $1/2$\} by evaluations, to leverage information propagation across clusters for feature enhancement. Totally, we train the network in an end-to-end manner for 80 epochs using the ADAM optimizer~\cite{Kingma2015AdamAM} and One-cycle~\cite{Smith2019SuperconvergenceVF} $lr$ scheduler with a maximum of 0.01 on a single NVIDIA GeForce RTX3090 GPU. 

As shown in Tab.~\ref{tab:kittitest}, we follow the official metric to compute $AP_{3D}$ score recalling 40 positions with IoU thresholds set to $0.7$, $0.5$, and $0.5$, respectively for
\textit{Car}, \textit{Pedestrian}, and \textit{Cyclist}, under Moderate difficulty level. Our proposed Shift-SSD achieves outstanding detection performance among methods of the same point-based genre, with the best results in \textit{Car} that even surpass several point-voxel and voxel-based counterparts. Specifically, we outperform the SOTA efficient method IA-SSD~\cite{zhang2022not} and DBQ-SSD~\cite{2022dbqssd} in \textit{Car} by $1.52\%$ and $2.26\%$, and in \textit{Cyclist} by $1.19\%$ and $0.33\%$, respectively.

\noindent
\textbf{Waymo}
\label{subsection:waymoext}
dataset is larger in scale compared to KITTI and contains 12$\mathrm{M}$ 3D annotations distributed in 200$\mathrm{K}$ 360-degree LiDAR samples with higher point density, capturing more complex scenes. Therefore, we follow~\cite{zhang2022not, 2022dbqssd} to quadruple the number of sampled points after each SSA module, i.e. $16384$ $\rightarrow$ $4096$ $\rightarrow$ $2048$ $\rightarrow$ $1024$, while keeping the remaining network configuration unchanged. We then train our Shift-SSD for $30$ epochs using similar optimization settings as for KITTI on $4$ NVIDIA GeForce RTX3090 GPU.

To evaluate, we compute two official metrics $\textrm{mAP}$ and $\textrm{mAPH}$ both with IoU threshold of $0.7$ for \textit{Vehicle} and $0.5$ for \textit{Pedestrian} and \textit{Cyclist}, under two difficulty levels. Results shown in Tab.~\ref{tab:waymo_val} highlight the superiority of our Shift-SSD regardless of categories,
levels, and metrics. Pointedly, we outperform IA-SSD~\cite{zhang2022not} by $3.59\%$, $3.05\%$, and $0.83\%$ in $\textrm{mAP}$, and $3.87\%$, $3.82\%$, and $1.35\%$ in $\textrm{mAPH}$, averaging by difficulty levels for \textit{Vehicle}, \textit{Pedestrian}, and \textit{Cyclist} respectively. Similar performance gains can also be observed compared to DBQ-SSD~\cite{2022dbqssd}. These advancements indicate that the more complex the input pointclouds, the better our Shift-SSD performs.   

\begin{figure*}[t!]
   \centering
    \includegraphics[width=1.0\linewidth]{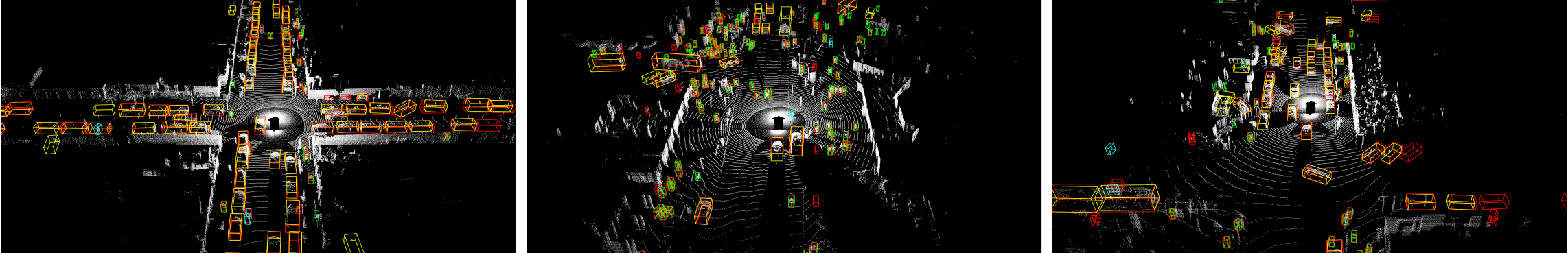}
    \caption{Qualitative results achieved by Shift-SSD on the \textit{validation} set of the Waymo Open Dataset~\cite{sun2020scalability}. Note that the \textit{Ground-truth} bounding boxes are shown in {\color{red}red}, the detected \textit{Vehicles'} are in {\color{yellow}yellow}, the \textit{Pedestrians'} are in {\color{green}green}, and the \textit{Cyclists'} are in {\color{cyan}cyan}.}
   \label{fig:WaymoValVis}

\end{figure*}

\noindent
\textbf{nuScenes} is another large-scale dataset providing 1.4$\mathrm{M}$ annotated 3D boxes for 40$\mathrm{K}$ keyframes and 390$\mathrm{K}$ LiDAR sweeps but has a more diverse set of 10 object categories compared to Waymo and KITTI. We adopt the same pointcloud preprocessing from~\cite{yang20203dssd} and adjust the prediction head to detect 10 classes while keeping the rest of network configuration unchanged as in Waymo experiments. Our Shift-SSD is then trained for 20 epochs using similar optimization and environment settings as in Waymo. 

As shown in Tab.~\ref{tab:nuscenesvalmAP}, we follow the official metric to compute $AP_{3D}$ score for each object category and the overall $mAP$. Our Shift-SSD obtains the best overall $mAP$ score of $44.39\%$. We respectively surpass
other point-based correlates including IA-SSD~\cite{zhang2022not} and 3DSSD~\cite{yang20203dssd} by substantial margins of
$2.16\%$ and $1.73\%$. Particularly, we largely exceed the performance of 3DSSD in the majority of classes such as \textit{Bicycle} ($8.06\%$), \textit{Bus} ($6.38\%$), \textit{Construction Vehicle} ($5.05\%$), and \textit{Trailer} ($3.56\%$). For the baseline IA-SSD, we outperform its results in all categories; for instance, \textit{Ped.} ($4.44\%$), \textit{Motor} ($2.99\%$), and \textit{Barrier} ($2.73\%$). These further highlight the performance of our Shift-SSD when dealing with challenging cases.    

\noindent
\textbf{Visualization}
Qualitative detection results achieved by Shift-SSD on the Waymo~\cite{sun2020scalability} dataset are selected and shown in Fig.~\ref{fig:WaymoValVis} for illustration.

\subsection{Ablation Studies}
\label{subsection:ablation}
This section reports the ablation study conducted on the KITTI dataset~\cite{geiger2013vision}. To reflect the performance under different difficulty levels, we evaluate the model by $AP_{Multi | Easy}$, $AP_{Multi | Mod.}$, and $AP_{Multi | Hard}$, defined by taking the average $\textrm{mAP}$ across classes under Easy, Moderate, and Hard difficulties, respectively.

\noindent
\textbf{Hyper-parameters Study.} As shown in Tab.~\ref{table:shiftingratio}, we evaluate the varied shifting proportions of channels and find that shifting $1/8$ ratio works best for fusing the local spatial features with the shifting features from the neighbor.

\begin{table}[t]
    \caption{Ablation study on different shifting ratios.\label{table:shiftingratio}}
    \centering
   \resizebox{0.45\textwidth}{!}{
    \renewcommand{\arraystretch}{1.05}
    \begin{tabular}{c|c|c|c|c|c}
    \Xhline{2.0\arrayrulewidth}
        Ratio  & 0 & 1/16 & 1/8 & 1/4 & 1/2 \\
    \Xhline{2.0\arrayrulewidth}
        AP    & 69.08 & 69.07 & \textbf{70.27} & 68.72 &69.08 \\
        \Xhline{2.0\arrayrulewidth}
    \end{tabular}
    }

\end{table}

\begin{table}[t!]
\centering
   \caption{Ablation study on different strategies to select the neighbor cluster to apply cross-cluster shifting.\label{tab:ClusterSelect}}
   \resizebox{0.47\textwidth}{!}{   
    \renewcommand{\arraystretch}{1.05}

   \begin{tabular}{p{18mm}<{\centering} | p{20mm}<{\centering} | p{20mm}<{\centering} | p{20mm}<{\centering}}
 
\Xhline{2.0\arrayrulewidth}
   \multirow{1}{=}{\centering Selection}
   & \multirow{1}{=}{$AP_{Multi | Easy}$}
   & \multirow{1}{=}{$AP_{Multi | Mod.}$}
   & \multirow{1}{=}{$AP_{Multi | Hard}$} \\

\Xhline{2.0\arrayrulewidth}
    Feats Scale & 77.75 \footnotesize{\color{BrickRed}(-1.71)}
                & 67.95 \footnotesize{\color{BrickRed}(-2.32)}
                & 64.76 \footnotesize{\color{BrickRed}(-2.28)} \\

    Nearest     & 78.54 \footnotesize{\color{BrickRed}(-0.92)}
                & 69.49 \footnotesize{\color{BrickRed}(-0.78)}
                & 65.84 \footnotesize{\color{BrickRed}(-1.20)} \\
                
   Points Num  & 78.96 \footnotesize{\color{BrickRed}(-0.50)}
               & 69.88 \footnotesize{\color{BrickRed}(-0.39)}
               & 66.78 \footnotesize{\color{BrickRed}(-0.26)} \\
\Xhline{2.0\arrayrulewidth}        
   Farthest    & \textbf{79.46}
               & \textbf{70.27}
               & \textbf{67.03}  \\

   \Xhline{2.0\arrayrulewidth}
   \end{tabular}} 

   \vspace{-0.4cm}
\end{table}

\noindent
\textbf{Selection Strategies on Clusters for Shifting.} As described in Section~\ref{subsection:CCS}, we measure the importance of neighbor clusters for applying cross-cluster shifting based on distance. Specifically, we select the farthest neighbor from the point set sampled within a ball region of the centering cluster point. We also consider different strategies other than Farthest Neighbor Sampling (denoted as Farthest in Tab.~\ref{tab:ClusterSelect}) in selecting the cluster to apply Cross-Cluster Shifting. As a counterpart of Farthest Neighbor Sampling, we experiment with Nearest Neighbor Sampling, denoted as Nearest in Tab.~\ref{tab:ClusterSelect}. Besides, we consider making a selection on neighbors to apply Cross-Cluster Shifting based on the scale of the features (calculated by taking mean along channel dimension), denoted as the ``Feats Scale in Tab.~\ref{tab:ClusterSelect}." Another way to measure the cluster importance is based on the number of points that the neighbor clusters are summarized from. As shown in Tab.~\ref{tab:ClusterSelect}, we empirically found that selecting the farthest cluster to apply for cross-cluster shifting leads to the best performance.

\noindent
\textbf{Effectiveness Study.} As shown in Tab.~\ref{tab:effectStudy}, we evaluate the effectiveness of our proposed Cross-Cluster Shifting (CS) against different exchanging schemes including concatenation (Concat), averaging (Avg), single-head attention (Attn), and no exchanging (None). It is noted that we use the same network architecture as our Shift-SSD excluding the exchanging design and employ the same neighbor cluster selection strategy across experiments to ensure a fair comparison. Our CS achieves the best results, especially in challenging cases of Moderate and Hard.

\noindent
\textbf{Efficiency Study.}
It is shown in Tab.~\ref{tab:efficiencyAbl} that our Shift-SSD is still lightweight and maintains a competitive efficiency among point-based detectors.

\begin{table}[t!]
\centering
   \caption{Ablation study on different exchanging operations.}
   \resizebox{0.47\textwidth}{!}{   
   \renewcommand{\arraystretch}{1.05}
   \begin{tabular}{c|c|c|c}
\Xhline{2.0\arrayrulewidth}

     \multirow{1}{*}{Ex. Op.}
   & \multirow{1}{*}{$AP_{Multi | Easy}$}
   & \multirow{1}{*}{$AP_{Multi | Mod.}$}
   & \multirow{1}{*}{$AP_{Multi | Hard}$}  \\
   
\Xhline{2.0\arrayrulewidth}
    None     & 79.13 \footnotesize{\color{BrickRed}(-0.33)}
         & 69.08 \footnotesize{\color{BrickRed}(-1.19)}
         & 65.51  \footnotesize{\color{BrickRed}(-1.52)} \\
    Concat & 78.16 \footnotesize{\color{BrickRed}(-1.30)} 
         & 68.91 \footnotesize{\color{BrickRed}(-1.36)} 
         & 65.11 \footnotesize{\color{BrickRed}(-1.92)}  \\
    Avg & 79.25 \footnotesize{\color{BrickRed}(-0.21)}
         & 69.08 \footnotesize{\color{BrickRed}(-1.19)}
         & 65.32 \footnotesize{\color{BrickRed}(-1.71)} \\
    Attn & 78.83 \footnotesize{\color{BrickRed}(-0.63)}
         & 68.67 \footnotesize{\color{BrickRed}(-1.60)}
         & 65.26 \footnotesize{\color{BrickRed}(-1.77)} \\

\Xhline{2.0\arrayrulewidth}
    \textbf{CS} & \textbf{79.46} 
              & \textbf{70.27}
              & \textbf{67.03}  \\
   \Xhline{2.0\arrayrulewidth}
   \end{tabular}}

   \label{tab:effectStudy}
   \vspace{0.1em}
\end{table}

\begin{table}[t]
    \caption{Efficiency evaluation. The number of learnable parameters is denoted as ``\#Params''.\label{tab:efficiencyAbl}}
    \centering
    \scalebox{1.0}{
    \renewcommand{\arraystretch}{1.05}
    \begin{tabular}{r|c|c}
    \Xhline{2.0\arrayrulewidth}
        Method   & Latency (ms) & \#Params
        (M) \\
    \Xhline{2.0\arrayrulewidth}
        PointRCNN \cite{shi2019pointrcnn}    & 98.04           &  4.04  \\
        3DSSD \cite{yang20203dssd}    & 90.91           &  2.51 \\
        IA-SSD \cite{zhang2022not}    & 43.48           & 2.70   \\
    \Xhline{2.0\arrayrulewidth}
        \textbf{Shift-SSD (Ours)} & \textbf{46.72}  & \textbf{2.78}\\
        \Xhline{2.0\arrayrulewidth}
    \end{tabular}
    }
    \vspace{-0.4cm}
\end{table}

\section{Conclusion}
We have presented a novel point-based 3D detector that incorporates our proposed efficient {\em Cross-Cluster Shifting} module. This approach not only boosts efficiency but also enhances accuracy through its cross-cluster modeling capability. Diverging from traditional point-based object detectors, which extract features within confined regions, our {\em Cross-Cluster Shifting} facilitates seamless information exchange between clusters. Leveraging these advancements, our Shift-SSD outperforms its peers in terms of performance while retaining optimal inference efficiency among all existing point-based 3D methods.

\noindent
\textbf{Acknowledgement}
The authors are thankful for the financial support from the Hetao Shenzhen-HongKong Science and Technology Innovation Cooperation Zone (HZQB-KCZYZ-2021055), this work was also supported by Shenzhen Deeproute.ai Co., Ltd (HZQB-KCZYZ-2021055).

\clearpage

{\small
\bibliographystyle{IEEEtran}
\bibliography{ref}
}

\end{document}